\documentclass[letterpaper, 10 pt, conference]{ieeeconf}  

\IEEEoverridecommandlockouts                              

\usepackage{url}
\usepackage{graphicx}
\usepackage{listings}
\usepackage{float}
\usepackage{flushend}
\usepackage{hyperref}

\usepackage{booktabs}
\DeclareFixedFont{\ttb}{T1}{txtt}{bx}{n}{12} 
\DeclareFixedFont{\ttm}{T1}{txtt}{m}{n}{12}  

\usepackage{color}
\definecolor{deepblue}{rgb}{0,0,0.5}
\definecolor{deepred}{rgb}{0.6,0,0}
\definecolor{deepgreen}{rgb}{0,0.5,0}

\newcommand\pythonstyle{\lstset{
language=Python,
basicstyle=\scriptsize,
morekeywords={self},              
keywordstyle=\ttb\color{deepblue},
emph={MyClass,__init__},          
emphstyle=\ttb\color{deepred},    
stringstyle=\color{deepgreen},
frame=tb,                         
showstringspaces=false
}}
\lstnewenvironment{python}[1][]
{
\pythonstyle
\lstset{#1}
}
{}

\newcommand\pythoninline[1]{{\pythonstyle\lstinline!#1!}}

\title{\LARGE \textbf{Integration of Large Language Models within Cognitive Architectures for Planning and Reasoning in Autonomous Robots}}

\author{Miguel Á. González-Santamarta$^{*,1}$, Irene González-Fernández, Francisco J. Rodríguez-Lera$^{1}$,\\Ángel Manuel Guerrero-Higueras$^{1}$, and Vicente Matellán-Olivera$^{1}$
\thanks{$^{*}${Correspondence to: \tt\small mgons@unileon.es}}%
\thanks{$^{1}$Robotics Group, University of León, Spain}%
}

\begin{document}

\maketitle
\thispagestyle{empty}
\pagestyle{empty}

\begin{abstract}
Symbolic reasoning systems have been used in cognitive architectures to provide inference and planning capabilities. However, defining domains and problems has proven difficult and prone to errors. Moreover, Large Language Models (LLMs) have emerged as tools to process natural language for different tasks. In this paper, we propose the use of LLMs to tackle these problems. This way, this paper proposes the integration of LLMs in the ROS 2-integrated cognitive architecture MERLIN2 for autonomous robots. Specifically, we present the design, development and deployment of how to leverage the reasoning capabilities of LLMs inside the deliberative processes of MERLIN2. As a result, the deliberative system is updated from a PDDL-based planner system to a natural language planning system. This proposal is evaluated quantitatively and qualitatively, measuring the impact of incorporating the LLMs in the cognitive architecture. Results show that a classical approach achieves better performance, but the proposed solution provides an enhanced interaction through natural language.
\end{abstract}

\section{Introduction}

Symbolic reasoning systems have long served as integral components in cognitive architectures for robotics, offering capabilities in inference and planning. Nevertheless, symbolic systems rely on predefined rules, leading to difficulties with complexity and adaptability. In contrast, the advent of Large Language Models (LLMs) has introduced new avenues for natural language processing across various tasks. These models, exemplified by their application in problem-solving contexts~\cite{Chalvatzaki2023,silver2023generalized,10161317,10161534,gragera2023exploring}, leverage extensive textual resources to tackle complex problems effectively, offering access to a diverse range of information and guidance.

Therefore, using LLMs in robotics can bring several benefits and capabilities. They have been used for natural language interaction \cite{koubaa2023rosgpt,palnitkar2023chatsim}, enabling robots to understand and generate human language, making it easier for users to communicate with and control robots with natural language. It also simplifies the knowledge retrieval~\cite{li2023modelscope,jiang2020x,linzbach2023decoding} since LLMs have vast knowledge repositories, which can be used by robots to access information, answer questions, or provide explanations to users. Additionally, they can also be used in explainability and interpretability~\cite{10.1007/978-3-031-40725-3_45} for robotics, leveraging the narrative capabilities of the models to explain and interpret the logs generated by robots.

LLMs have recently achieved leading performance in tasks involving arithmetic and reasoning using the method of prompting known as chain-of-thought (CoT)~\cite{wei2022chain}, which encourages complex multi-step reasoning by providing step-by-step answer examples. Studies such as \cite{kojima2023large} recently achieved significant performance in such tasks using LLMs that are capable of zero-shot reasoning when it is added the instruction \textit{Let us consider this step by step} in the prompt.

Behavior generation in autonomous robots is still an open problem that has been faced by different paradigms. Within them, we can find the deliberative architectures~\cite{ingrand2017deliberation} that use planning approaches, the subsumption architectures~\cite{brooks1991new}, and the reactive architectures~\cite{peter1997experiences}. The combination of these paradigms produces hybrid architectures~\cite{arkin1997aura}, which significantly impacts the robotics community.

Our proposal focuses on reasoning and presenting the integration of LLMs in our existing cognitive architecture. The proposed approach seeks to update the current deliberative module, which is based on a symbolic planner, 
by one based on LLMs. Therefore, this paper analyzes the use of llama\_ros, available in \cite{llama_ros_2023,gonzálezsantamarta2025integratingquantizedllmsrobotics}, in the MERLIN2~\cite{GONZALEZSANTMARTA2023100477,González-Santamarta2024}, cognitive architecture. It allows using offline LLMs inside robotics systems. Thus, its use is proposed as an alternative to classic deliberative models based on symbolic knowledge expressed in PDDL. Finally, a quantitative and qualitative review of its effect and impact on the overall decision-making system is also described in the paper.

\textbf{Contribution:} The main contribution of this research is the contextualization and evaluation of using an LLM as a reasoning system inside a cognitive architecture. 


In robotics, reasoning~\cite{levesque2008cognitive} is the capability of logically and systematically processing the knowledge of the robot. There are several types of reasoning \cite{ye2018survey}, for instance, practical reasoning or planning, to find the next best action and perform it; and theoretical reasoning, which aims at establishing or evaluating beliefs. The reasoning approach that is faced in this work is closer to a planning type, generating sequences of actions to achieve the robot's goals; and evaluating whether a specific goal is already achieved.

The evaluation of this proposal has been made in a human-robot interaction environment carried out in a mock-up apartment using a service robot. The reasoning faced in these missions consists of planning and evaluating whether a specific condition will be satisfied following the execution of a sequence of actions that commences from an initial state. 
Thus, we initially present this proposal as the implications of using LLMs in cognitive architectures. In other words, LLMs can be used in service robots, although some could not see them as a formal reasoning mechanism. 

The rest of the paper is organized as follows. Section~\ref{sec:sota} reviews the state of the art and the background. Section~\ref{sec:material} depicts the material and methods of this work, that is, the llama\_ros tool, the cognitive architecture MERLIN2, the integration process, and the hardware and software setup. Section~\ref{sec:eval} presents the evaluation of this work, composed of the experiment setup, the results and the discussions. Finally, Section~\ref{sec:conclusions} presents the conclusions and the future work.

\section{Background and Related Work}
\label{sec:sota}
In recent years, months, and weeks, not only the scientific community has seen remarkable progress in the field of generative AI and large language models (LLMs) but also the general public,  generating a perception that it could be used everywhere. This section will review LLMs and their use for inference in current models. 

\textbf{LLMs.} A Large Language Model (LLM) is an artificial intelligence system trained on vast amounts of text data to understand and generate human language. The release of chatGPT~\cite{openai2023gpt4} (GPT-3.5 and GPT-4). However, the release of LLaMA~\cite{llama1} and LLaMA2~\cite{llama2}, in their different sizes, 7B, 13B, 33B, and 65B; has marked a new age of LLMs since it allows researchers to train models on custom datasets.

Nevertheless, LLMs require substantial computational resources for inference and deployment. However, managing the computational burden is a significant challenge in resource-constrained environments like embedded systems within robots. Quantization~\cite{wu2020integer,cai2017deep,lin2015neural,lin2016fixed}, in this context, refers to the process of reducing the precision of the model's parameters and activations, typically from floating-point numbers to fixed-point numbers. By doing so, quantization dramatically reduces LLMs' memory and computational requirements, making them feasible for deployment in robots with limited processing power and memory. It is essential since it enables robots to leverage the power of LLMs for natural language understanding, decision-making, and interaction while operating efficiently within their constrained hardware environments. 

The LLaMA models and the quantization methods allow the proliferation of a significant number of LLMs that can be deployed in personal computers and embedded systems with tools like llama.cpp~\cite{githubGitHubGgerganovllamacpp}. LLMs continue to evolve rapidly with the introduction of innovative models such as Alpaca~\cite{alpaca}, Vicuna~\cite{vicuna}, WizardLM~\cite{wizardlm}, Nous-Hermes~\cite{NousHermes}, and Marcoroni~\cite{marcoroni}, which contribute to the growing arsenal of robust language understanding and generation tools.

\textbf{Deliberative and reasoning capabilities of LLMs.} As LLMs have grown increasingly sophisticated and capable, their ability to engage in meaningful deliberation and planning has become a research subject. Deliberation and planning entail carefully considering various options, arguments, and perspectives before deciding and strategically organizing actions or steps to achieve a particular goal. Some works have attempted to use LLMs as planners. For instance, \cite{guan2023leveraging} uses PDDL for planning, while \cite{song2023llmplanner} explores using few-shot planning in embodied agents like robots.

Despite pre-trained models being widely recognized for their remarkable few-shot learning abilities in various natural language processing tasks, a recent prompting technique called chain-of-thought (CoT)~\cite{wei2022chain} has achieved state-of-the-art performance. In \cite{kojima2023large}, it has been proved that LLMs can also excel as zero-shot reasoners. This technique has been expanded by applying a search algorithm for better results. The tree-of-thought~\cite{yao2023tree} allows LLMs to perform deliberate decision-making by considering different reasoning paths, self-evaluating them, and deciding the next course of action. Another case is the graph-of-thought~\cite{besta2023graph} that is similar to the previous case but distributes the possible paths in a graph format instead of a tree.

In robotics, we can find more works that tried to perform PDDL planning with pre-trained LLMs~\cite{silver2022pddl}. More advanced research like ProgPrompt~\cite{singh2023progprompt} enables plan generation through a programmatic LLM prompt structure. However, LLMs are rarely used within cognitive architectures.

\textbf{Cognitive Architectures.} Cognitive architectures serve as the foundational framework for autonomous robots, guiding their perception, decision-making, and action execution. These architectures can be broadly categorized into several classes, each offering unique advantages and characteristics tailored to specific robotic applications. Its use allows us to understand the relationship between the knowledge, the perception, and the action of such a robot.
A taxonomy of cognitive architectures is posed in the literature~\cite{ye2018survey,kotseruba202040}. There are three categories: symbolic architectures, similar to deliberative architectures; emergent architectures, which replace reactive architectures and emphasize the connectionist concept; and hybrid architectures.

The most extended cognitive architecture category is the hybrid approach. For instance, HiMoP hybrid architecture is proposed in our previous works \cite{rodriguez2018himop}. It is composed of a deliberative system that uses PDDL (Planning Domain Definition Language)~\cite{PDDL} to represent the knowledge of the robot; a reactive system with a pool of state machines; and a motivational system, that contains all the robot needs. The world state of the robot is defined using the PDDL, which is the common language to carry out planning in robotics, while the state machines are used to implement the actions that the robot can perform.
Alternatively, there are different technical approaches such as  MOBAR~\cite{munoz2019mobar} and CORTEX~\cite{bustos2019cortex}, where the knowledge of the robot is represented by a knowledge graph that holds symbolic and geometric information within the same structure or Ginés et al.~\cite{gines2022depicting,martin2021client}, where the reactive system is based on behavior trees~\cite{colledanchise2021implementation}. These are the architectures that have guided the development of MERLIN2~\cite{GONZALEZSANTMARTA2023100477,González-Santamarta2024}, the one used in this paper.


\begin{figure*}[ht]
\centering
\vspace{2mm}
\includegraphics[width=0.97\textwidth]{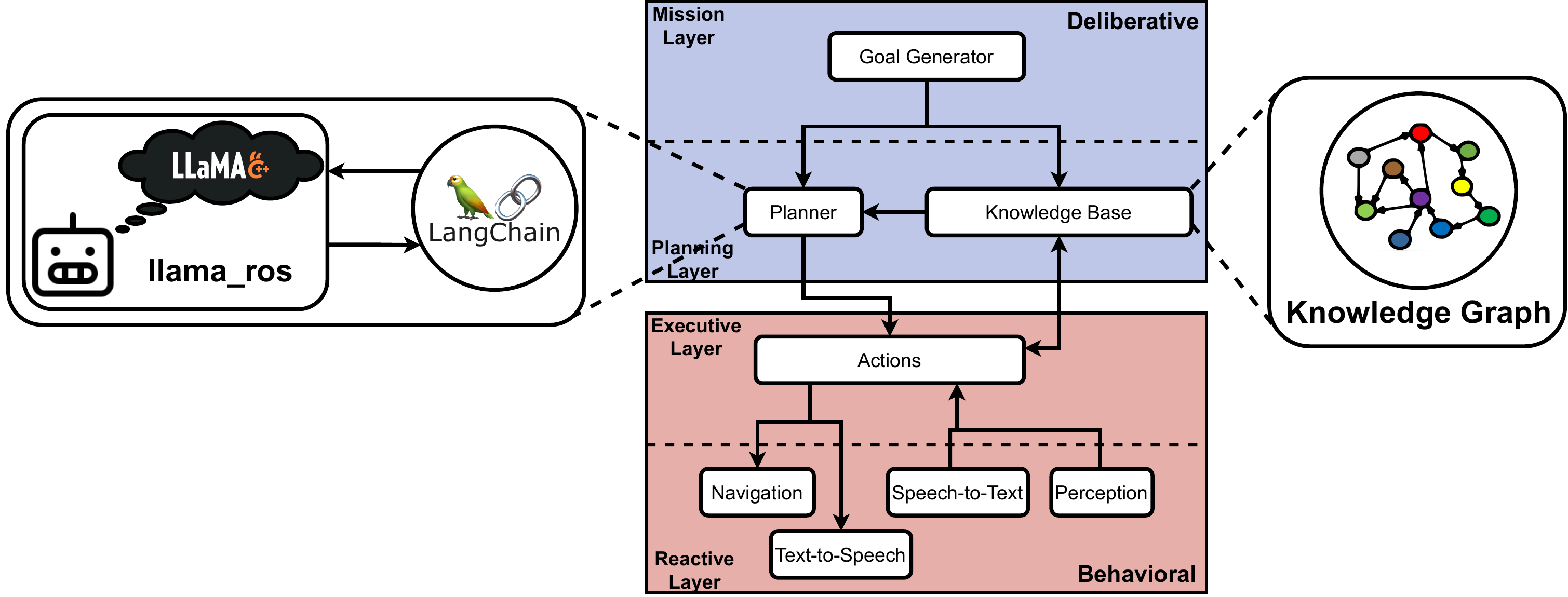}
\caption{MERLIN2 architecture diagram. The original architecture is formed by the Deliberative System and the Behavioral System, which are divided into four layers: the Mission Layer, Planning Layer, Executive Layer and Reactive Layer. LLMs are integrated into the Planning Layer by using llama\_ros and LangChain. Besides, a knowledge graph replaces the original knowledge base.\label{fig:merlin2} } 
\end{figure*}

\section{Materials and Methods}
\label{sec:material}

This section lays the groundwork for the detailed integration of LLMs into the cognitive architecture known as MERLIN2. We delve into the llama\_ros tool, which allows for integrating  LLMs into ROS 2 \cite{macenski2022robot}. Additionally, we provide an overview of the MERLIN2 architecture, discuss how LLMs are integrated into MERLIN2, and explain how these models enhance the reasoning capabilities within the system. We also cover the technical setup, including both hardware and software components.

\subsection{llama\_ros}

The llama\_ros tool~\cite{llama_ros_2023}, available in a public repository\footnote{\url{https://github.com/mgonzs13/llama_ros}}, integrates llama.cpp~\cite{githubGitHubGgerganovllamacpp} into ROS 2 through a suite of packages. It enables the execution of LLaMA-based models and other LLMs on devices with low computing performance using integer quantization. This is achieved through a C/C++ implementation, facilitating model deployment across various platforms, including GPU support.



The llama\_ros tool introduces a ROS 2 node, offering primary LLM functionalities such as:

\begin{itemize}
    \item \textit{Response generation}: similar to applications like ChatGPT, llama\_ros can generate responses to prompts from humans or other ROS 2 nodes. This is done using a ROS 2 action server.

    \item \textit{Tokenize}: a ROS 2 service for text tokenization, essential for the model's language processing. Tokens, which may represent characters, words, or other textual elements, are processed into sequences for the model. 

    \item \textit{Embeddings}: another ROS 2 service produces text embeddings. These embeddings represent tokens in a high-dimensional space, each dimension capturing different linguistic features. They are critical for understanding and distinguishing language components.
\end{itemize}

These interfaces are particularly valuable for advanced prompt engineering methods~\cite{sahoo2024systematic}, such as using the embedding service to transform text into vectors for a database. This database supports Retrieval Augmented Generation (RAG) \cite{lewis2020retrieval} by allowing the retrieval of vectors related to specific texts, leading to the creation of more precise prompts.


Furthermore, to enhance prompt engineering, llama\_ros has been incorporated into LangChain~\cite{Chase_LangChain_2022}, a platform that eases the development of LLM applications. Integrating the aforementioned interfaces with LangChain enables the utilization of this framework within ROS 2 environments.

\subsection{MERLIN2}

Cognitive architectures enable robots to execute complex behaviors. In this research, we use MERLIN2~\cite{GONZALEZSANTMARTA2023100477,González-Santamarta2024}, a hybrid cognitive architecture integrated into ROS 2 and designed for autonomous robots, which is accessible in a public repository\footnote{\url{https://github.com/MERLIN2-ARCH/merlin2}}. Figure~\ref{fig:merlin2} depicts MERLIN2. Drawing on methodologies from existing literature \cite{rodriguez2018himop,munoz2019mobar,bustos2019cortex,gines2022depicting}, MERLIN2 incorporates symbolic knowledge to model the robot's world, a deliberative system for planning to achieve goals, state machines for immediate behaviors, and emergent modules for object recognition, and both speech recognition and synthesis. These elements split into two systems: the Deliberative and the Behavioural Systems. 

\begin{figure}[ht]
\centering
\vspace{2mm}
\includegraphics[width=0.34\textwidth]{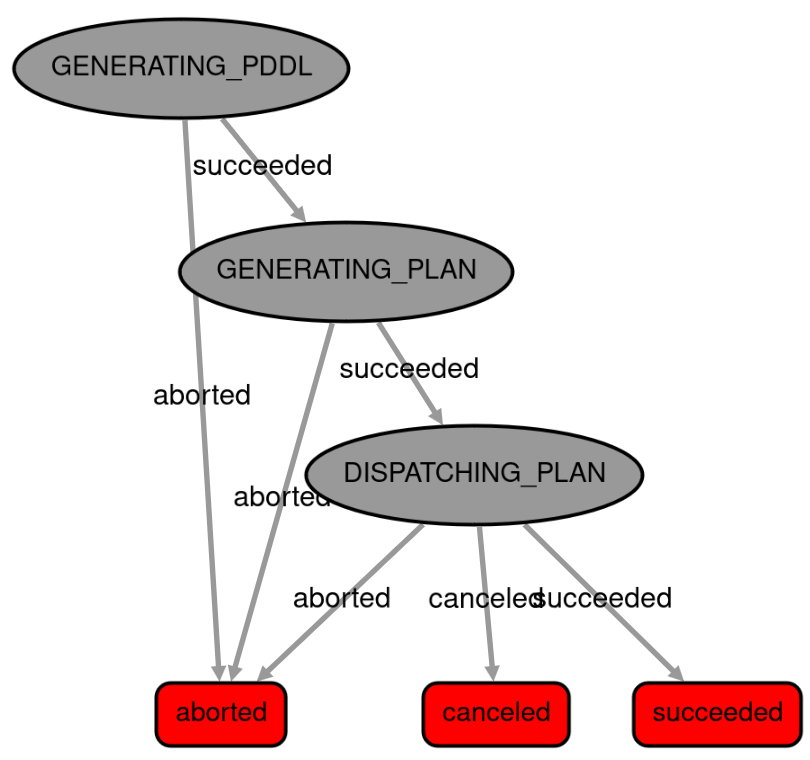}
\caption{YASMIN state machine in charge of controlling the original Planning Layer of MERLIN2. It has three states: one to generate the PDDL from the knowledge base, another state to generate the plans using PDDL planners and another state to execute the plan.\label{fig:executor_fsm} } 
\end{figure}

The Deliberative System manages high-level tasks and gathers the Mission Layer and the Planning Layer. The Mission Layer sets high-level objectives for the robot, using a state machine to generate these goals. The Planning Layer, based on traditional deliberative approaches, maintains a symbolic knowledge base in PDDL format to create action plans that fulfill the robot's goals, employing established PDDL planners like POPF~\cite{popf}. Control over the Planning Layer is exercised through a state machine built with YASMIN~\cite{yasmin} (shown in Figure~\ref{fig:executor_fsm}). This machine is in charge of generating PDDL from the knowledge base, crafting the plan with the planner, and executing the planned actions.

The Behavioural System collects another two layers: the Executive and the Reactive Layers. The Executive Layer orchestrates the robot's immediate actions, leveraging its skills for short-term behaviors. These actions can be structured using either state machines or behavior trees for organization and execution. On the other hand, the Reactive Layer consolidates the robot's skills, encompassing capabilities like navigation, text-to-speech, and speech-to-text recognition.




\subsection{Integrating LLMs into MERLIN2}

Incorporating LLMs into MERLIN2 aims to use their advanced reasoning abilities, transforming the existing Planning Layer. This modification involves substituting the current symbolic elements, specifically the PDDL planner and the knowledge base, as illustrated in Figure~\ref{fig:merlin2}, with LLM-driven components.

\begin{figure}[ht]
\centering
\vspace{2mm}
\includegraphics[width=0.44\textwidth]{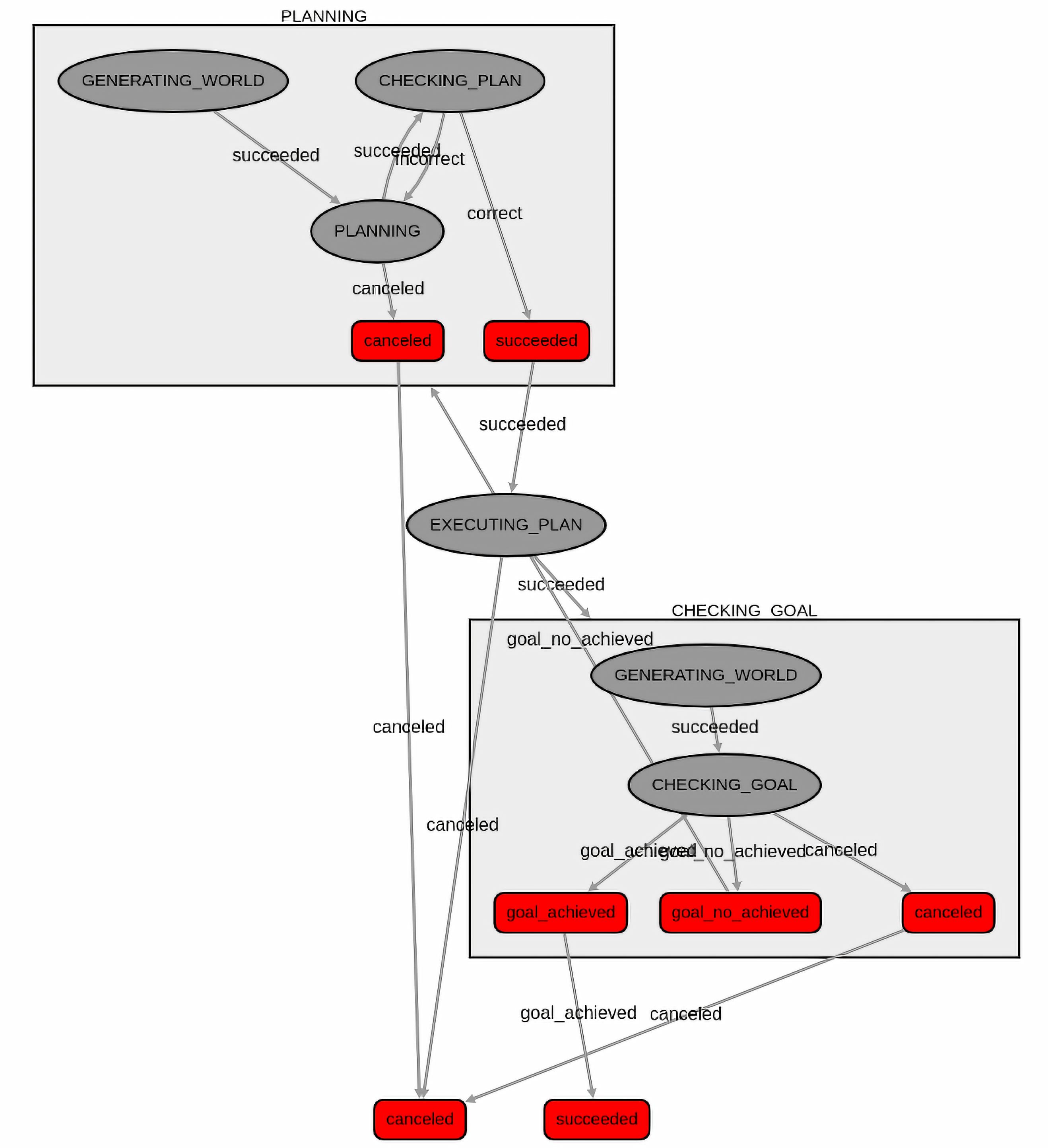}
\caption{YASMIN state machine in charge of controlling the new Planning Layer obtained after integrating the LLMs into MERLIN2. It has a nested state machine to generate plans for the goals, a state to execute the plan, and another nested state machine to check if the goal is achieved. \label{fig:agi4ros_fsm}}
\end{figure}

The redefined layer is managed by a YASMIN state machine, depicted in Figure~\ref{fig:agi4ros_fsm}. This setup features two nested state machines -- PLANNING and CHECKING\_GOAL -- and a singular state, EXECUTING\_PLAN. Within this structure, one nested state machine is dedicated to generating plans to meet the robot's goals, another to verify the accomplishment of these goals, and a separate state is in charge of the execution of the plans.

\subsubsection{Knowledge Graph}

The knowledge base in MERLIN2 is now replaced with a knowledge graph, accessible at a public repository\footnote{\url{https://github.com/mgonzs13/knowledge_graph}}, building on approaches from prior studies \cite{bustos2019cortex,gines2022depicting,martin2021client}. This shift enhances how knowledge is organized through the graph's structure. Furthermore, employing a knowledge graph to encapsulate the robot's knowledge not only refines the cognitive architecture's practicality but also facilitates quick comprehension of the robot's knowledge base by human operators.

\subsubsection{Generating the World State}

The world state is an intermediary form of knowledge representation, derived from the robot's knowledge graph by translating graph data into discrete knowledge items. This form is particularly suited for inclusion in the prompts directed towards the LLM.

The structure of the knowledge items is detailed as follows: nodes are represented by the format ``node is a type (properties)'' and edges adhere to the schema ``node relationship node (properties)''. The ``(properties)'' segment contains key-value pairs encapsulating the information associated with a node or an edge, such as waypoint coordinates.

Incorporating RAG enhances the creation of the world state, converting knowledge items into vectors and storing them within a vector database. Then, retrieval techniques are employed to selectively access knowledge items related to the robot's current goal. This process uses ChromaDB~\cite{trychromaAInativeOpensource}, a high-performance in-memory vector database, for efficient management and retrieval of vectorized knowledge items.

\begin{figure}[!ht]
\centering
\vspace{2mm}
\includegraphics[width=0.48\textwidth]{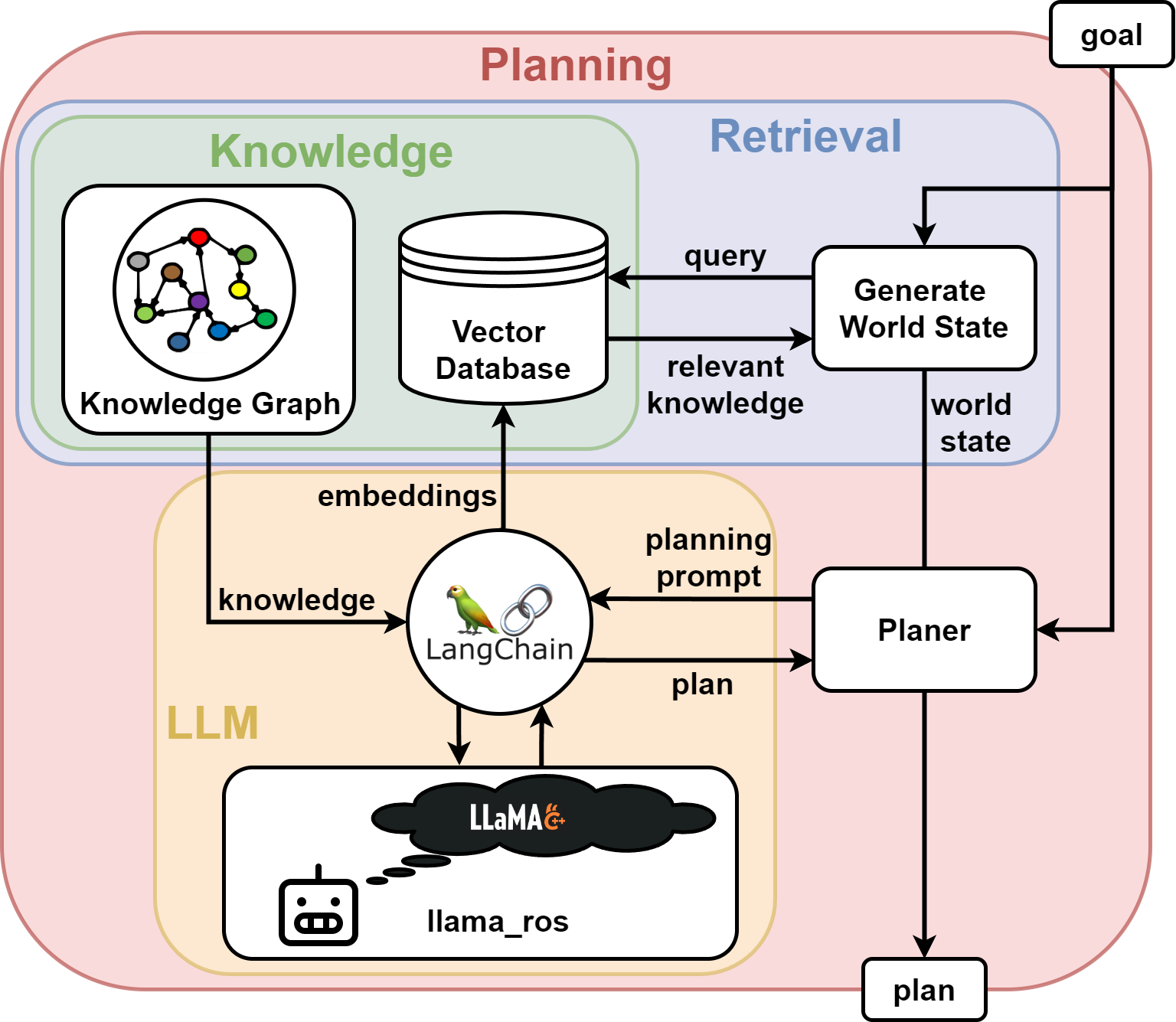}
\caption{Pipeline to perform the planning using LLMs and RAG inside the resulting cognitive architecture. First, the knowledge of the robot is converted into embeddings employing the LLM. Using the goal, a query is created to retrieve the relevant knowledge, that is, the world state. Then, the world state and the goal are used to create the planning prompt, which is used to prompt the LLM, through LangChain, to generate the plan.\label{fig:llm_planning} } 
\end{figure}

\subsubsection{Planning}
The PLANNING state machine creates plans addressing the robot's objectives, aligning with the original function of generating PDDL plans. The entire planning process utilizing LLMs is illustrated in Figure~\ref{fig:llm_planning}.

First, the robot's world state is compiled through RAG, selectively gathering knowledge pertinent to the goal. This world state, alongside the goal, forms the basis of a planning prompt that enables the LLM to function as a planner. The prompt encapsulates the robot's possible actions, its current world state, and its goal.

By using zero-shot CoT \cite{kojima2023large}, the LLM is then engaged via LangChain and llama\_ros to formulate a plan that meets the specified goal. Additionally, a grammar employing Backus-Naur Form (BNF) \cite{o2003grammatical} constrains the LLM's output to JSON\footnote{\url{https://github.com/ggerganov/llama.cpp/blob/master/grammars/json.gbnf}}, simplifying parsing efforts. Lastly, the plan's validity is assessed by verifying the accuracy of the response format.


\subsubsection{Executing the Plan}

After the plan's creation, the EXECUTING\_PLAN state, analogous to the original state, DISPATCHING\_PLAN, designated for executing formulated plans, runs the actions delineated by the LLM. The execution of each action has the potential to update the knowledge graph with its outcomes.

\subsubsection{Checking Goal}

After the plan is executed, the CHECKING\_GOAL state machine initiates. Similar to the process of the PLANNING state machine, it generates the world state and engages the LLM via LangChain by using its zero-shot reasoning abilities to assess if the goal has been met. The prompt for verifying goal achievement gathers both the current world state and the robot's intended goal. Additionally, this state generates a rationale, detailing whether the goal has been fulfilled or not.

Should the LLM conclude that the objective remains unattained, the entire state machine will reinitiate the planning process, this time incorporating the rationale behind the unmet goal to refine and enhance the subsequent plan generation.

\subsection{Hardware and Software Setup}

For the experiments conducted in this study, we utilized a computer outfitted with an Intel(R) i9-13980HX processor, 32 GB of RAM, and an RTX 4070 Nvidia graphics card. The operating system and framework of choice were Ubuntu 22.04 and ROS 2 Humble, respectively. The llama\_ros component was set up to employ a pretrained 4-bit quantized 13B Marcoroni LLM, optimized for GPU performance via cuBLAS. This setup involved activating 30 out of the total 43 layers for processing.

\section{Evaluation}
\label{sec:eval}

This section presents the evaluation of MERLIN2 after integrating LLMs. It is based on comparing the resulting architecture with its original version MERLIN2. First, the experiment setup and the metrics are presented. Then, the results and the discussion are presented.

\begin{table*}[ht!]
	\centering
        \vspace{2mm}
	\caption{Results for the first experiment with 6 missions.}
	\label{tab:mission_1}
	{
		\begin{tabular}{lrrrrrrrrrr}
			\toprule
			\multicolumn{1}{c}{} & \multicolumn{5}{c}{Execution Time (Seconds)} & \multicolumn{5}{c}{Traveled Distance (Meters)} \\
			\cline{2-6}\cline{7-11}
			 & MERLIN2 & FI & NRI & NCI & NRNCI & MERLIN2 & FI & NRI & NCI & NRNCI  \\
			\cmidrule[0.4pt]{1-11}
			Mean & $107.728$ & $251.625$ & $178.476$ & $200.149$ & $166.469$ & $30.579$ & $15.297$ & $17.254$ & $14.540$ & $20.797$  \\
			Std. Deviation & $12.023$ & $14.403$ & $18.054$ & $23.336$ & $11.173$ & $1.023$ & $5.853$ & $7.678$ & $5.787$ & $3.518$  \\
			Minimum & $99.316$ & $237.411$ & $159.253$ & $165.308$ & $150.256$ & $29.879$ & $10.763$ & $8.841$ & $5.569$ & $16.748$  \\
			Maximum & $128.349$ & $270.434$ & $197.291$ & $228.392$ & $179.281$ & $32.368$ & $23.238$ & $26.580$ & $20.462$ & $25.106$  \\
			Sum & $538.642$ & $1258.126$ & $892.378$ & $1000.747$ & $832.345$ & $152.895$ & $76.483$ & $86.272$ & $72.700$ & $103.987$  \\
			\bottomrule
		\end{tabular}
	}
\end{table*}

\begin{table*}[ht!]
	\centering
	\caption{Results for the second experiment with 20 missions.}
	\label{tab:mission_2}
	{
		\begin{tabular}{lrrrrrrrrrr}
			\toprule
			\multicolumn{1}{c}{} & \multicolumn{5}{c}{Execution Time (Seconds)} & \multicolumn{5}{c}{Traveled Distance (Meters)} \\
			\cline{2-6}\cline{7-11}
			 & MERLIN2 & FI & NRI & NCI & NRNCI & MERLIN2 & FI & NEV & NCI & NRNCI  \\
			\cmidrule[0.4pt]{1-11}
			Mean & $257.763$ & $848.196$ & $563.292$ & $665.627$ & $508.021$ & $62.957$ & $53.100$ & $51.723$ & $49.373$ & $53.701$  \\
			Std. Deviation & $65.448$ & $50.052$ & $20.817$ & $26.855$ & $35.405$ & $27.258$ & $13.334$ & $8.648$ & $10.894$ & $10.434$  \\
			Minimum & $162.720$ & $800.366$ & $540.865$ & $626.202$ & $471.818$ & $23.564$ & $36.653$ & $41.260$ & $34.059$ & $40.753$  \\
			Maximum & $339.039$ & $929.419$ & $588.901$ & $691.266$ & $556.861$ & $100.373$ & $72.377$ & $61.253$ & $60.445$ & $64.954$  \\
			Sum & $1288.815$ & $4240.980$ & $2816.460$ & $3328.134$ & $2540.106$ & $314.786$ & $265.500$ & $258.614$ & $246.867$ & $268.503$  \\
			\bottomrule
		\end{tabular}
	}
\end{table*}

\subsection{Experimental Setup}

\begin{figure}[ht]
\centering
\includegraphics[width=0.243\textwidth]{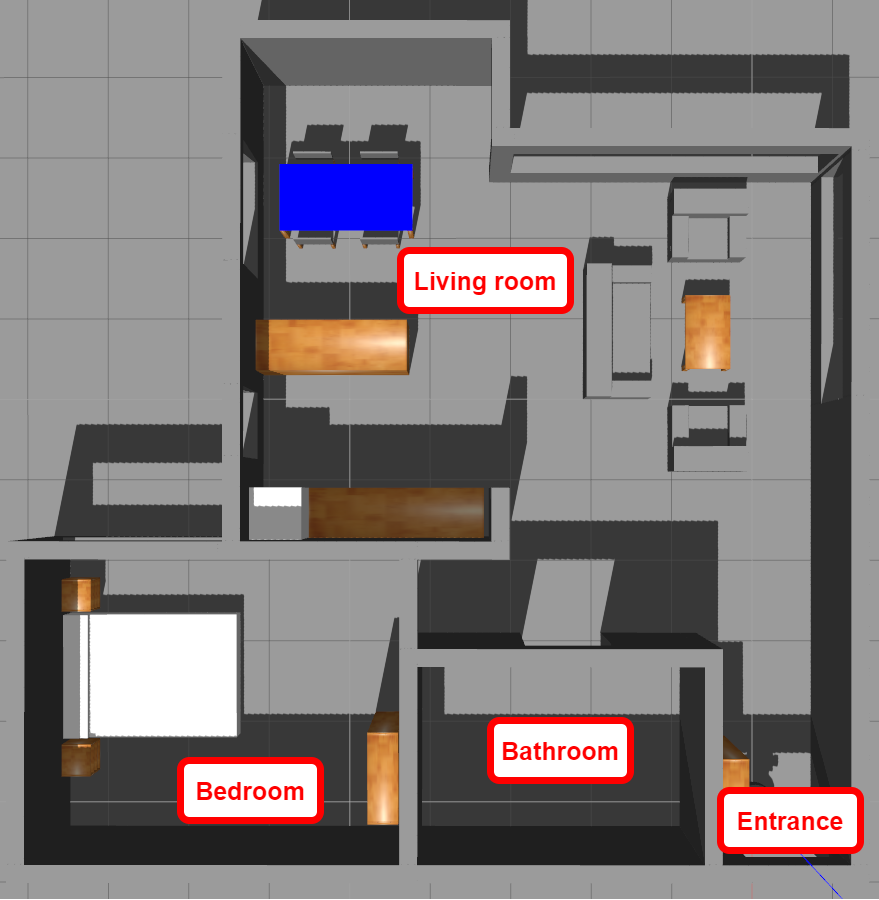}
\includegraphics[width=0.22\textwidth]{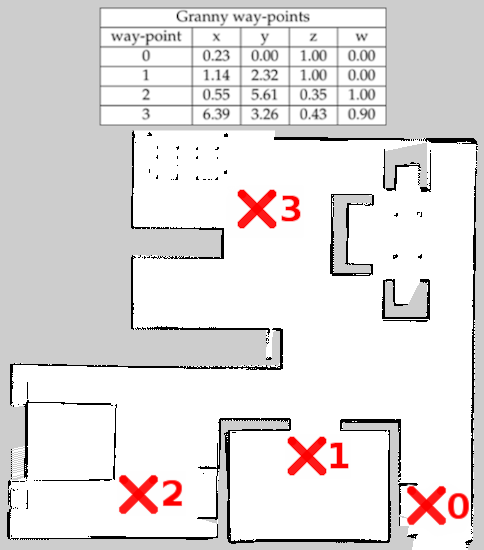}
\caption{Gazebo zenith view and map deployed for the experimental validation of the resulting cognitive architecture after integrating LLMs. The map shows the four points that are used to perform the experiment. These points contain a person that the robot must greet in each iteration of the experiment.}
\label{fig:granny_map}
\end{figure}

The evaluation method is similar to the one depicted in MERLIN~\cite{gonzalez2020merlin}. Thus, in this work, the experiments consist of greeting people in the rooms of an apartment using the RB1 robot, which is a service robot composed of a differential mobile base, with a LiDAR, and a torso, with an RGB-D camera and speakers. As a result, the robot must be able to navigate through the apartment and speak to people. Figure~\ref{fig:granny_map} shows the simulated apartment and the navigation map reused in this experiment. Moreover, the metrics used are the execution time, in seconds, required by the robot to complete the missions; and the traveled distance, in meters, covered by the robot during the execution of the missions.

Based on the presented missions, two experiments have been designed. Both of them are based on greeting people missions, but they differ in the number of missions. For each experiment, a randomized list of people that the robot needs to greet is created. Besides, half of the missions are canceled after 10 seconds of starting to evaluate the impact of switching missions. Hence, the first experiment comprises 6 missions, meaning greeting 6 times, of which 3 of them are canceled. The second involves 20 missions, that is, greeting 20 times, of which 10 are canceled.



For these experiments, the actions implemented for the experiments are the navigation and greeting actions. The navigation action uses Nav2~\cite{macenski2020marathon2} to move the robot between the waypoints. Thus, this action receives as arguments: the robot name, the source waypoint and the target waypoint. The greeting action uses a text-to-speak tool to greet the four people. Its arguments are the robot's name, the person's name and the waypoint of the person.

\begin{figure}[ht]
\centering
\includegraphics[width=0.46\textwidth]{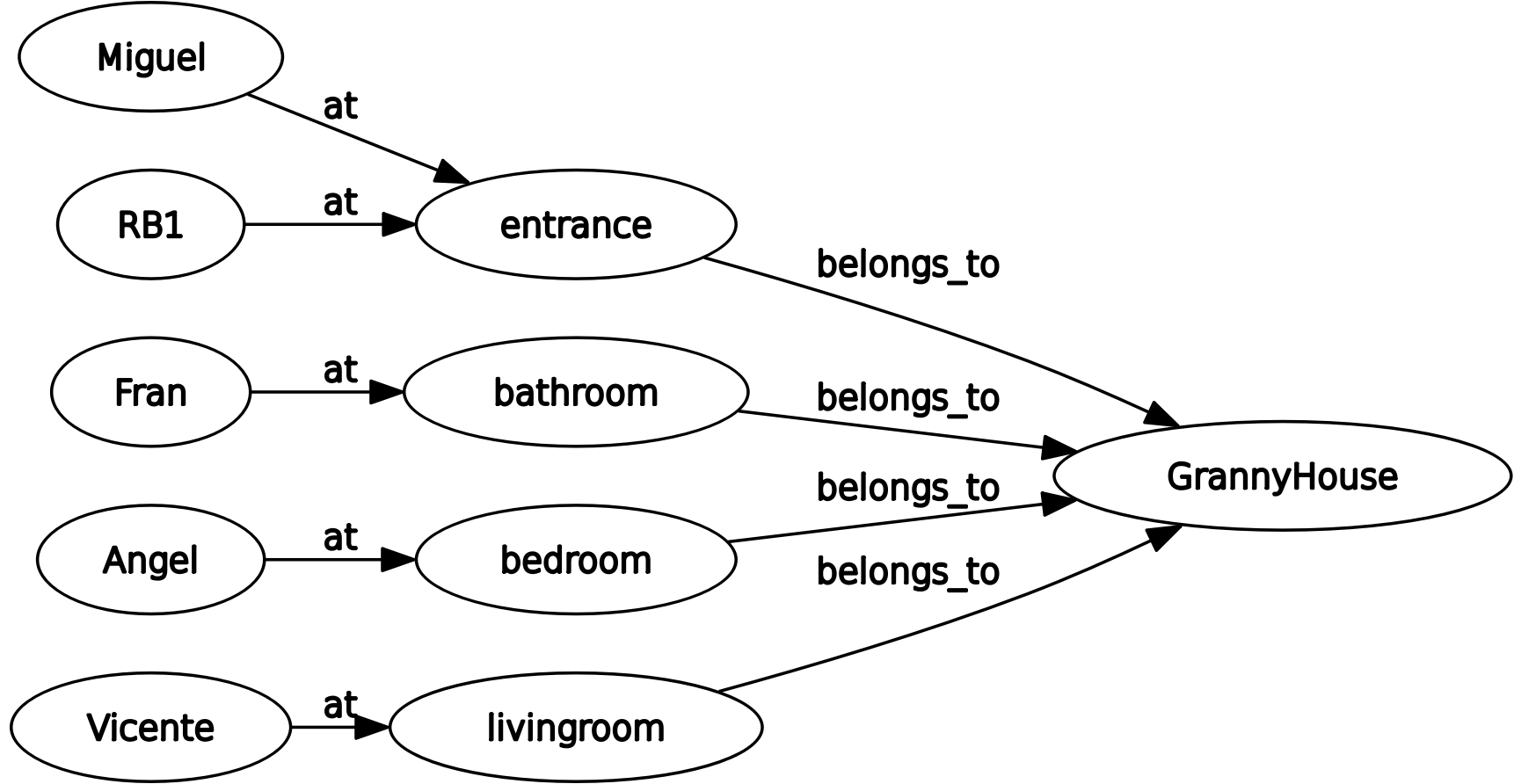}
\caption{Knowledge graph for the experiments. It shows an apartment named GrannyHouse with four rooms, which are the entrance, the bathroom, the bedroom and the living room. There is a person in each room: Miguel at the entrance, Fran in the bathroom, Angel in the bedroom and Vicente in the living room. Finally, the robot RB1 is at the entrance.\label{fig:kg} } 
\end{figure}

On the other hand, the initial knowledge graph for the experiment is presented in Figure~\ref{fig:kg}. It shows that the apartment is named GrannyHouse and has four rooms, which corresponds with the marked waypoints in the map of Figure~\ref{fig:granny_map}. The rooms are the entrance, bathroom, bedroom and living room. Besides, there are four people in different rooms, as shown in the graph. Finally, the robot RB1 is at the entrance.

Finally, the two experiments are executed with the original architecture MERLIN2 and with four different versions of the architecture with the LLMs integration:
\begin{itemize}
    \item Full Integration (FI): This version uses all the components mentioned before in the integration part. The world state is generated with RAG and a vector database, ChromaDB, and the goal is checked with the CHECKING\_GOAL state. The number of retried knowledge items from the vector database is 10.
    \item No RAG Integration (NRI): This version is similar to the Full Version but does not use a vector database. As a result, it uses all the knowledge items from the graph as the world states that there are 19 knowledge items.
    \item No CHECKING\_GOAL State Integration (NCI): This version is similar to the Full Version but without the CHECKING\_GOAL State. As a result, the architecture depends only on the planning capability of the LLM.
    \item No RAG and No CHECKING\_GOAL State Integration (NRNCI): This version uses neither RAG nor the CHECKING\_GOAL state.
\end{itemize}

\subsection{Results}
\label{sec:results}

Table~\ref{tab:mission_1} shows the results for the first experiment, where each execution is composed of 6 goals. MERLIN2 presents the lowest mean time value of 107.728 seconds, while FI presents the most significant time value of 251.625 s. NRI, NCI and NRNCI get time values of 178.476 s, 200.149 s and 166.469 s. On the other hand, MERLIN2 has the highest value of traveled distance, 30.579 m, while the others have similar values of 15.297 m, 17.254 m, 14.540  and 20.797 m.

Table~\ref{tab:mission_2} shows the results for the second experiment, where each execution is composed of 20 goals. In this experiment, MERLIN2 also presents the lowest mean time value of 257.763 seconds, while FI presents the highest time value of 848.196 s. NRI, NCI and NRNCI obtained time values of 563.292 s, 665.627 s and 508.021 s. On the other hand, MERLIN2 has the most significant value of traveled distance, 62.957 m, while the others have similar values of 53.100 m, 51.723 m, 49.373 m and 53.701 m.

\subsection{Discussion}
\label{sec:discussion}

The applicability of using an LLM for reasoning processes is evident. The experiment showcases a robot navigating and moving between points, similarly to the process conducted with the traditional PDDL framework.

Yet, when assessing both the time taken and the distance traversed during the mission, the performance under the traditional PDDL framework significantly outshines that of the LLM approach. In the initial experiment, the MERLIN2 architecture demonstrated a performance that was 2.3 times higher than the FI and 1.5 times better than the NRNCI version. Moreover, the outcomes of the second experiment reinforce MERLIN2 as the preferable option for decision-making over the LLM alternatives, being 3.2 times more efficient compared to FI and 1.9 times more effective against NRNCI. Conversely, when comparing NRI and NCI, NCI's execution time is 1.2 times slower than NRI in both experiments. This discrepancy highlights that employing RAG incurs a higher cost than merely using the LLM to verify goal achievement.

The advantage is also noticeable regarding the distance covered. This parameter allows for measuring effective service duration, as longer distances entail greater battery consumption. On average, outcomes based on the LLM depicted a scenario where the robot covered merely half the distance compared to operating under MERLIN2. This discrepancy stems from the LLM versions requiring more time for the planning phase than MERLIN2.


Finally, throughout the development and integration process, it has been observed that quantized LLMs tend to exhibit a higher sensitivity to prompts compared to their unquantized counterparts. Additionally, several challenges were identified with incorporating LLMs within the cognitive architecture:

\begin{itemize}
    \item Increased World State Complexity: for complex environments, the world state -- and consequently, the prompt size -- may expand. Employing RAG can mitigate this issue by streamlining the prompt size, albeit at the cost of extended execution times.
    
    \item Feedback from CHECKING\_GOAL: although it may prolong the execution period, the feedback obtained from the CHECKING\_GOAL process can significantly benefit the planning stage, particularly when re-planning is required by unmet goals. However, it further prolongs the overall execution timeframe.
\end{itemize}


\section{Conclusions}
\label{sec:conclusions}

We have presented the integration of LLMs into MERLIN2. Leveraging the LLM's reasoning capabilities, a new Planning Layer has been built and experimentation has been carried out to evaluate the integration process in human-robot interaction missions within a simulated environment.

It is worth discussing the PDDL performance revealed by the results of our experiments, which is superior to LLM in its different approaches. However, part of the community would prefer to interact in a more natural language manner, as is done with the LLM options, which is achieved thanks to the quantized LLMs and the llama\_ros tool integrated into MERLIN2. As a result, the cognitive architecture can present an improved human-robot interaction degree.

In future works, we propose to improve the performance by utilizing smaller and more accurate LLMs and specific embedding models to improve the RAG process. Besides, we will explore the use of graph algorithms to try to optimize the world state instead of using RAG, as well as the use of other symbolic techniques like anthologies. Finally, we want to include the new Visual Language Models (VLM), which can be used to extract natural language text from the images captured by the robot. These ongoing efforts can contribute to the evolution and refinement of cognitive architectures, incorporating cutting-edge language models.



   









\section*{ACKNOWLEDGMENT}
This work has been partially funded by an FPU fellowship provided by the Spanish Ministry of Universities (FPU21/01438); by the grant PID2021-126592OB-C21 funded by MCIN/AEI/10.13039/501100011033; and by the Recovery, Transformation, and Resilience Plan, financed by the European Union (Next Generation) thanks to the TESCAC project (Traceability and Explainability in Autonomous Systems for improved Cybersecurity) granted by INCIBE to the University of León.



 \bibliographystyle{unsrt}
 \bibliography{root}

\end{document}